\documentclass[journal]{IEEEtran}
\usepackage{graphicx}
\usepackage{amsmath}
\usepackage{amssymb}
\usepackage{color}
\usepackage{multirow}
\usepackage[letterpaper=true,colorlinks,bookmarks=false]{hyperref}
\usepackage{subfigure}
\usepackage[lined,boxed, ruled]{algorithm2e}
\usepackage{multirow}
\usepackage{url}

\begin{document}

\title{CE-Net: Context Encoder Network for 2D Medical  Image Segmentation}

\author{Zaiwang Gu, Jun Cheng, Huazhu Fu, Kang Zhou, Huaying Hao, Yitian Zhao, Tianyang Zhang, Shenghua Gao and Jiang Liu
\thanks{This work is supported in part by Cixi Institute of Biomedical Engineering, Chinese Academy of Sciences (Y60001RA01, Y80002RA01), Ningbo 3315 Innovation team grant, Zhejiang Province (Y61102DL03),  National Natural Science Foundation of China (61601029); and also the Zhejiang Provincial Natural Science Foundation of China (LZ19F010001)). \it{(Corresponding Author : Jun Cheng)}}
\thanks{Z.~Gu is with School of Mechatronic Engineering and Automation, Shanghai University, Shanghai 200072 and also with Cixi Institute of Biomedical Engineering, Chinese Academy of Sciences, Zhejiang 315201, China. Email: guzaiwang@nimte.ac.cn.}

\thanks{J.~Cheng,  H.~Hao, Y.~Zhao and T.~Zhang are with Cixi Institute of Biomedical Engineering, Chinese Academy of Sciences, Zhejiang 315201, China. Email: \{chengjun, haohuaying, yitian.zhao, zhangtianyang\}@nimte.ac.cn.}

\thanks{H. Fu is with the Inception Institute of Artiﬁcial Intelligence, Abu Dhabi, United Arab Emirates, and also with the Institute for Infocomm Research, Agency for Science, Technology and Research, Singapore 138632 (e-mail: huazhufu@gmail.com).}

\thanks{K.~Zhou and S.~Gao are  with School of Information Science and Technology, ShanghaiTech University, Shanghai 201210, China. Email: zhoukang@shanghaitech.edu.cn, gaoshh@shanghaitech.edu.cn.}

\thanks{J. Liu is  with the Department
of Computer Science and Engineering, Southern University of Science and
Technology, Guangdong 518055, China and also with the Cixi Institute of Biomedical Engineering, Chinese Academy of Sciences, Zhejiang 315201, China. Email: liuj@sustech.edu.cn.}

}
\markboth{}%
{Gu \MakeLowercase{\textit{et al.}}:}

\maketitle

\begin{abstract}
Medical image segmentation is an important step in medical image analysis. With the rapid development of convolutional neural network in image processing, deep learning has been used for medical image segmentation, such as optic disc segmentation, blood vessel detection, lung segmentation,  cell segmentation, etc. Previously, U-net based approaches have been proposed.  However, the consecutive pooling and strided convolutional operations lead to the loss of some spatial information.  In this paper, we propose a context encoder network (referred to as CE-Net) to capture more high-level information and preserve spatial information for 2D medical  image segmentation. CE-Net mainly contains three major components: a feature encoder module, a context extractor and a feature decoder module. We use pretrained ResNet block as the fixed feature extractor.  The context extractor module is formed by a newly proposed dense atrous convolution (DAC) block and residual multi-kernel pooling (RMP) block. We applied the proposed CE-Net to different 2D medical image segmentation tasks. Comprehensive results show that the proposed method outperforms the original U-Net method and other state-of-the-art methods for optic disc segmentation, vessel detection, lung segmentation, cell contour segmentation and retinal optical coherence tomography layer segmentation.
\end{abstract}

\begin{IEEEkeywords}
Medical image segmentation, Deep Learning, Context encoder network
\end{IEEEkeywords}

\IEEEpeerreviewmaketitle

\section{Introduction} Medical image segmentation is often an important step in medical image analysis, such as optic disc segmentation \cite{cheng2013superpixel, fu2018joint,  aquino2010detecting} and blood vessel detection \cite{fu2016deepvessel, zhao2015automated, liskowski2016segmenting, roychowdhury2015iterative, azzopardi2015trainable} in retinal images, cell segmentation  \cite{al2010improved,ronneberger2015u, song2017dual} in electron microscopic (EM) recordings, lung segmentation \cite{wang2017central,shen2016learning,song2016lung, lee2001automated, heimann2007statistical} and brain segmentation  \cite{makropoulos2014automatic,menze2015multimodal,cherukuri2018learning,huang2014brain,kumar2018infinet,chen2009automated} in computed tomography (CT) and magnetic resonance imaging (MRI).
Previous approaches to medical image segmentation are often based on edge detection and template matching \cite{lee2001automated}. For example, circular or elliptical Hough transform are used in optic disc segmentation \cite{zhu2008detection, aquino2010detecting}. Template matching is also used for spleen segmentation in MRI sequence images \cite{mihaylova2018spleen} and ventricular segmentation in brain CT images \cite{chen2009automated}.


Deformable models are also proposed for medical image segmentation. The shape-based method using level sets \cite{tsai2003shape} has been proposed for two-dimensional segmentation of cardiac MRI images and three-dimensional segmentation of prostate MRI images. In addition, a level set-based deformable model is adopted for kidney segmentation from abdominal CT images \cite{khalifa2011new}.
The deformable model has also been integrated with the Gibbs prior models for segmenting the boundaries of organs \cite{chen2000image}, with an evolutionary algorithm and a statistical shape model to segment the liver \cite{heimann2007statistical} from CT volumes.
In optic disc segmentation, different deformable models have also been proposed and adopted, such as mathematical morphology, global elliptical model, local deformable model \cite{lowell2004optic}, and modified active shape model \cite{xu2007optic}.

Learning based approaches are proposed to segment medical images as well. 
 Aganj \emph{et al.} \cite{aganj2018unsupervised} proposed the local center of mass based method for unsupervised learning based image segmentation in X-ray and MRI images. Kanimozhi \emph{et al.} \cite{kanimozhi2013brain} applied the stationary wavelet transform to obtain the feature vectors, and self-organizing map is adopted to handle these feature vectors for unsupervised MRI image segmentation. Tong \emph{et al.} \cite{tong2015discriminative} combined dictionary learning and sparse coding to segment multi-organ in abdominal CT images. Pixel classification based approaches \cite{abramoff2007automated}, \cite{cheng2013superpixel} are also learning based approaches which train classifiers based on pixels  using pre-annotated data.  However, it is not easy to select the pixels and extract features to train the classifier from the larger number of pixels. Cheng \emph{et al.} \cite{cheng2013superpixel} used the superpixel strategy to reduce the number of pixels and performed the optic disc and cup segmentation using superpixel classification.
Tian \emph{et al.} \cite{tian2016superpixel} adopted a superpixel-based graph cut method to segment 3D prostate MRI images. In \cite{kitrungrotsakul2015liver},  superpixel learning based method is integrated with restricted regions of shape constrains to segment lung from CT images.

The drawbacks of these methods lie in the utilization of hand-crafted features to obtain the segmentation results. On the one hand, it is difficult to design the representative features for different applications. On the other hand,  the designed features working well for one type of images often fail on another type. Therefore, there is a lack of general approach to extract the feature.

With the development of convolutional neural network (CNN) in image and video processing \cite{krizhevsky2012imagenet} and medical image analysis \cite{zhou2018multi, wang2017zoom}, automatic feature learning algorithms using deep learning have emerged as feasible approaches for medical image segmentation. Deep learning based segmentation methods are pixel-classification based learning approaches. Different from traditional pixel or superpixel classification approaches which often use hand-crafted features, deep learning approaches learn the features and overcome the limitation of hand-crafted features. 

Earlier deep learning approaches for  medical image segmentation are mostly based on image patches. Ciresan \emph{et al.} \cite{ciresan2012deep}  proposed to  segment neuronal membranes in microscopy images based on patches and sliding window strategy. Then, Kamnitsas \emph{et al.} \cite{kamnitsas2017efficient} employed a multi-scale 3D CNN architecture with fully connected conditional random field (CRF) for boosting patch based brain lesion segmentation.    Obviously, this solution introduces two main drawbacks: redundant computation caused from sliding window and the inability to learn global features.

With the emerging of the end-to-end fully convolutional network (FCN) \cite{long2015fully}, 
Ronneberger \emph{et al.} \cite{ronneberger2015u} proposed U-shape Net (U-Net) framework for biomedical image segmentation. U-Net  has shown promising results on the neuronal structures segmentation in electron microscopic recordings and cell segmentation in light microscopic images.
It has becomes a popular neural network architecture for biomedical image segmentation tasks \cite{norman2018use, sevastopolsky2017optic, roy2017relaynet, skourt2018lung}.  Sevastopolsky \emph{et al.} \cite{sevastopolsky2017optic} applied U-Net to directly segment the optic disc and optic cup in retinal fundus images for glaucoma diagnosis. Roy \emph{et al.} \cite{roy2017relaynet} used a similar network for retinal layer segmentation in optical coherence tomography (OCT) images. Norman \emph{et al.} \cite{norman2018use} used U-Net to segment cartilage and meniscus from knee MRI data. The U-Net is also applied to directly segment lung from CT images \cite{skourt2018lung}. 

Many variations have been made on U-Net  for different medical image segmentation tasks. Fu \emph{et al.} \cite{fu2016deepvessel} adopted the CRF to gather the multi-stage feature maps for boosting the vessel detection performance. 
Later, a modified U-Net framework (called M-Net) \cite{fu2018joint} is proposed for joint optic disc and cup segmentation by adding multi-scale inputs and deep supervision into the U-net architecture. Deep supervision mainly introduces the extra loss function associated with the middle-stage features. Based on the deep supervision, Chen \emph{et al.} \cite{chen2016voxresnet} proposed a Voxresnet to segment volumetric brain, and Dou \emph{et al.} \cite{dou20163d} proposed 3D deeply supervised network (3D DSN) to automatically segment lung in CT volumes. 

\begin{figure*}
	\centering
	\includegraphics[height = 9.5cm]{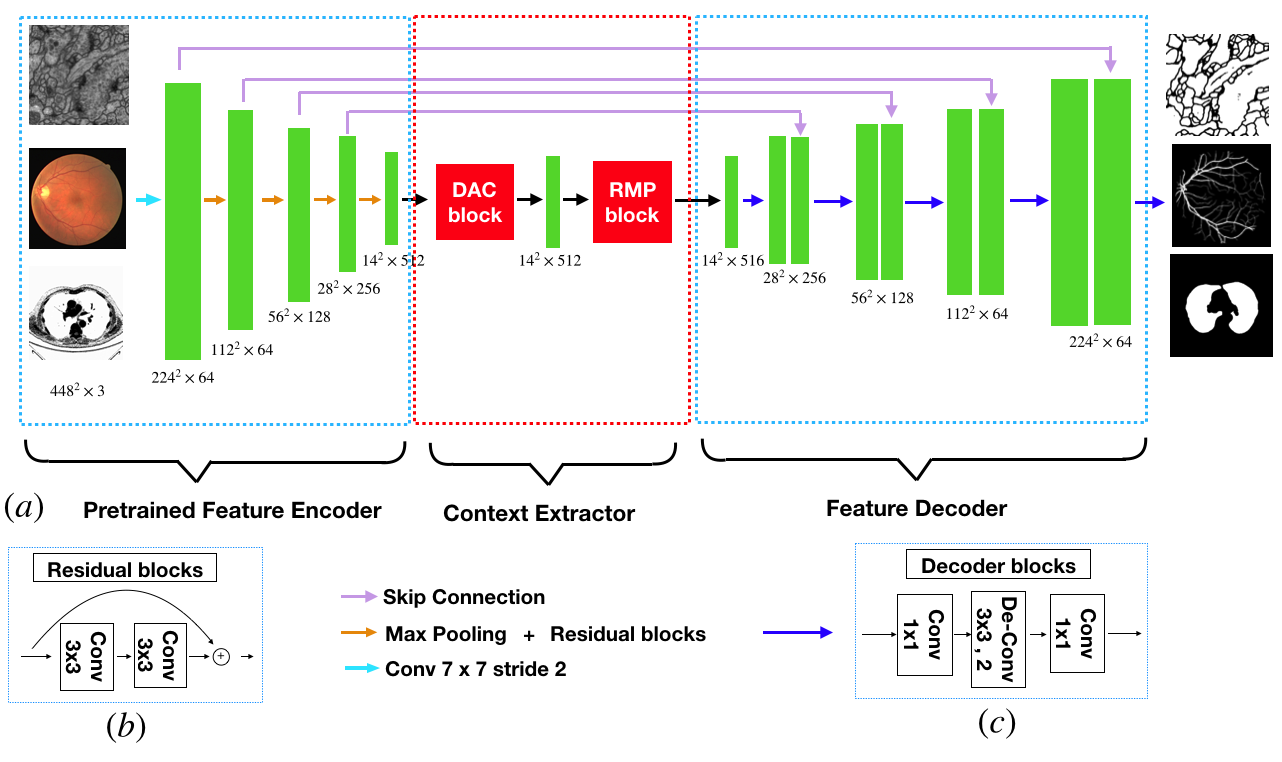}
	\caption{Illustration of the proposed CE-Net. Firstly, the images are fed into a feature encoder module, where the ResNet-34 block pretrained from ImageNet is used to replace the original U-Net encoder block.  The  context extractor is proposed to generate more high-level semantic feature maps. It contains a dense atrous convolution (DAC) block and a residual multi-kernel pooling (RMP) block.  Finally, the extracted features are fed into the feature decoder module. In this paper, we adopt a decoder block to enlarge the feature size, replacing the original up-sampling operation. The  decoder block  contains 1$\times$1 convolution and 3$\times$3 deconvolution operations. Based on skip connection and the decoder block, we  obtain the mask as the segmentation prediction map.}
	\label{2}
\end{figure*}

To enhance the feature learning ability of U-Net, some new  modules have been proposed to replace the original blocks. 
Stefanos \emph{et al.} \cite{apostolopoulos2017pathological} proposed a branch residual U-network (BRU-net) to segment pathological OCT retinal layer for age-related macular degeneration diagnosis. BRU-net relies on residual connection and dilated convolutions to enhance the final OCT retinal layer segmentation.  Gibson \emph{et al.} \cite{gibson2018automatic} introduced dense connection in each encoder block to automatically segment multiple organs on abdominal CT.  Kumar \emph{et al.} \cite{kumar2018infinet} proposed an InfiNet for infant brain MRI segmentation.
Besides the above achievements for  U-Net based medical image segmentation, some researchers have also  made progress to modify U-Net for   general image segmentation. Peng \emph{et al.} \cite{peng2017large} proposed a novel global convolutional network to improve semantic segmentation. Lin \emph{et al.} \cite{lin2017refinenet} proposed a multi-path refinement network, which contains residual convolution unit, multi-resolution fusion and chained residual pooling. Zhao \emph{et al.} \cite{zhao2017pyramid} adopted spatial pyramid pooling to gather the extracted feature maps to improve the semantic segmentation performance. 

A common limitation of the U-Net and its variations is that the consecutive pooling operations or convolution striding reduce the feature resolution to learn increasingly abstract feature representations. Although this invariance is beneficial for classification or object detection tasks, it often impedes dense prediction tasks which require detailed spatial information. Intuitively, maintaining high-resolution feature maps at the middle stages can boost segmentation performance. However, it increases the size of feature maps, which is not optimal to accelerate the training and ease the difficulty of optimization. Therefore, there is a trade-off between accelerating the training and maintaining the high resolution. Generally, the U-Net structures can be considered as Encoder-Decoder architecture. The Encoder aims to reduce the spatial dimension of feature maps gradually and capture more high-level semantic features. The Decoder aims to recover the object details and spatial dimension. Therefore,
it is spontaneous to capture more high-level features in the encoder and preserve more spatial information in the decoder to improve the performance of image segmentation.

 Motivated by the above discussions and also the Inception-ResNet structures \cite{he2016deep, szegedy2017inception} which make the neural network wider and deeper, we propose a novel dense atrous convolution (DAC) block to employ atrous convolution. The original U-Net architecture captures multi-scale features in the limited scaling range by adopting the consecutive 3$\times$3 convolution and pooling operations in the encoding path. Our proposed DAC block could capture wider and deeper semantic features by infusing four cascade branches with multi-scale atrous convolutions. In this module, the residual connection is utilized to prevent the gradient vanishing.
In addition, we also propose a residual multi-kernel pooling (RMP) motivated from spatial pyramid pooling \cite{he2014spatial}. The RMP block further encodes the multi-scale context features of the object extracted from the DAC module by employing various size pooling operations, without the extra learning weights.  In summary, the DAC block is proposed to extract enriched feature representations with multi-scale atrous convolutions, followed by the RMP block for further context information with multi-scale pooling operations. 
Integrating the newly proposed DAC block and the RMP block with the backbone encoder-decoder structure, we propose a novel context encoder network named as CE-Net. It relies on the DAC block and the RMP block to get more abstract features and preserve more spatial information to boost the performance of medical image segmentation.

The main contributions of this work are summarized as follows:
	
\begin{enumerate}

\item{We propose a DAC block and RMP block to capture more high-level features and preserve more spatial information.}

\item{We integrate the proposed DAC block and RMP block with encoder-decoder structure for medical image segmentation.  }

\item{We apply the proposed method in different tasks including optic disc segmentation, retinal vessel detection, lung segmentation, cell contour segmentation and retinal OCT layer segmentation. Results show that the proposed method outperforms the state-of-the-art methods in these different tasks.  }
\end{enumerate}

The remainder of this paper is organized as follows. Section~\ref{Method} introduces the proposed method in details. Section~\ref{Experiment} presents the experimental results and discussions. In Section~\ref{Conclusion}, we draw some conclusions.

\section{Method}
\label{Method}
The proposed CE-Net consists of three major parts: the feature encoder module, the context extractor module, and the feature decoder module, as shown in Fig. \ref{2}.  

\subsection{Feature Encoder Module}

In U-Net architecture, each block of encoder contains two convolution layers and one max pooling layer. In the proposed method, we replace it with the pretrained ResNet-34 \cite{he2016deep} in the feature encoder module, which retains the first four feature extracting blocks without the average pooling layer and the fully connected layers. Compared with the original block, ResNet adds shortcut mechanism to avoid the gradient vanishing and accelerate the network convergence, as shown in Fig.~\ref{2}(b). For convenience, we use the modified U-net with pretrained ResNet as ‘backbone’ approach. 
\subsection{Context Extractor Module}
The context extractor module is a newly proposed module, consisting of  the DAC block and  the RMP block. This module extracts context semantic information and generates more high-level feature maps.
\begin{figure}[h]
	\centering
	\subfigure{
	  \includegraphics[width=7.5cm]{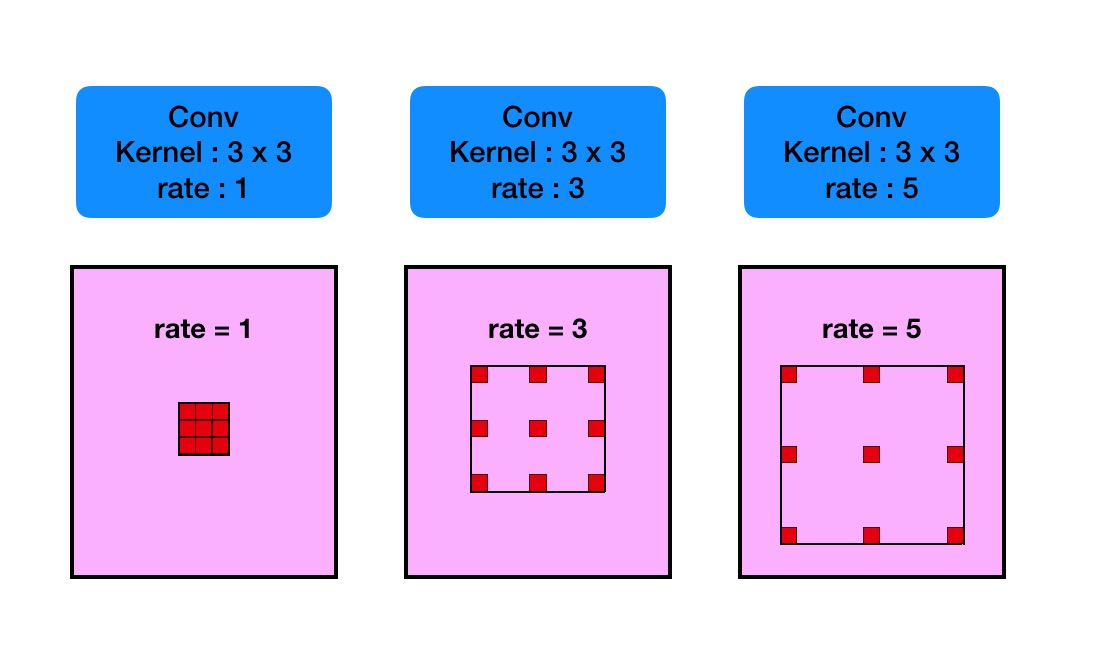}}	
	\caption{The illustrations of atrous convolution.}
	\label{picture1}
\end{figure}

1) \textbf{Atrous convolution}: In semantic segmentation tasks and object detection tasks, deep convolutional layers have shown to be effective in extracting feature representations for images. However, the pooling layers  lead to the loss of semantic information in images. In order to overcome this limitation, atrous convolution is adopted for dense segmentation \cite{chen2017deeplab}:

\begin{figure}[h]
	\centering
	\subfigure{
	  \includegraphics[width=8cm]{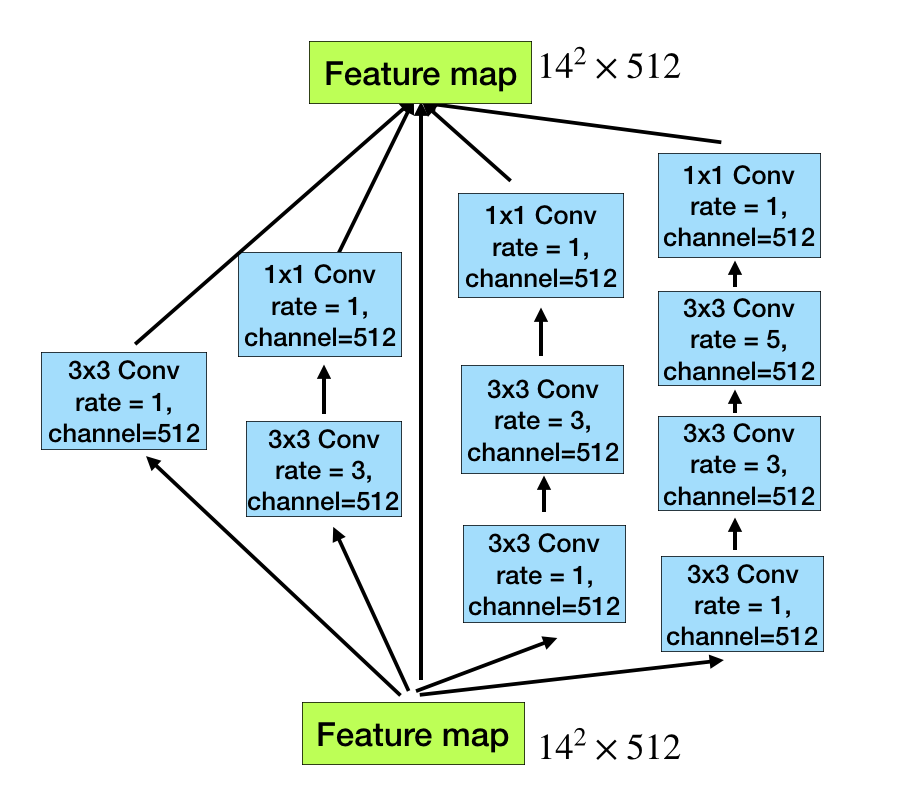}}
	\caption{The illustrations of dense atrous convolution block. It contains four cascade branches with the gradual increment of the number of atrous convolution, from 1 to 1, 3, and 5, then the receptive field of each branch will be 3, 7, 9, 19. Therefore, the network can extract features from different scales.}
	\label{picture2}
\end{figure}

The atrous convolution is originally proposed for the efficient computation of the wavelet transform. Mathematically, the atrous convolution under two-dimensional signals is computed as follows:
\begin{equation}
	\textbf{\emph{y}}[\textbf{\emph{i}}]=\sum_{\textbf{\emph{k}}} \textbf{\emph{x}}[\textbf{\emph{i}} + r \textbf{\emph{k}}]\textbf{\emph{w}}[\textbf{\emph{k}}],
\end{equation}
where the convolution of the input feature map \textbf{\emph{x}} and a filter \textbf{\emph{w}} yields the output \textbf{\emph{y}}, and the atrous rate {\emph{r}} corresponds to the stride with which we sample the input signal. It is equivalent to convolute
the input \textbf{\emph{x}} with upsampled filters produced by inserting
{\emph{r}} $-$ 1 zeros between two consecutive filter values along each
spatial dimension (hence the name atrous convolution in which
the French word atrous means holes in English). Standard
convolution is a special case for rate {\emph{r}} = 1, and atrous
convolution allows us to adaptively modify filter’s field-of-view by changing the rate value. See Fig.~\ref{picture1} for illustration.

2) \textbf{Dense Atrous Convolution module}: Inception\cite{szegedy2017inception} and ResNet\cite{he2016deep} are two classical and representative architectures in the  deep learning. Inception-series structures adopt different receptive fields to widen the architecture.
On the contrary, ResNet employs shortcut connection mechanism to avoid the exploding and vanishing gradients. It makes the neural network break through up to thousands of layers for the first time.  Inception-ResNet \cite{szegedy2017inception} block, which combines the Inception and ResNet, inherits the advantages of both approaches. Then it becomes a baseline approach in the field of deep CNNs. 

Motivated by the Inception-ResNet-V2 block and atrous convolution, we propose  dense atrous convolution (DAC) block to encode the high-level semantic feature maps. As shown in Fig.~\ref{picture2}, the atrous convolution is stacked in cascade mode. In this case, DAC has four cascade branches with the gradual increment of the number of atrous convolution, from 1 to 1, 3, and 5, then the receptive field of each branch will be 3, 7, 9, 19.  It employs different receptive fields, similar to Inception structures. In each atrous branch, we apply one 1$\times$1 convolution for rectified linear activation. Finally, we directly add the original features with other features, like shortcut mechanism in ResNet. Since the proposed block looks like a densely connected block, we name it dense atrous convolution block. Very often, the convolution of large reception field could extract and generate more abstract features for large objects, while the convolution of small reception field is better for small object. By combining   the atrous convolution of different atrous rates, the DAC block is able to extract features for  objects with various sizes.

\begin{figure}[!t]
	\centering
	\subfigure{
	  \includegraphics[width=7.8cm]{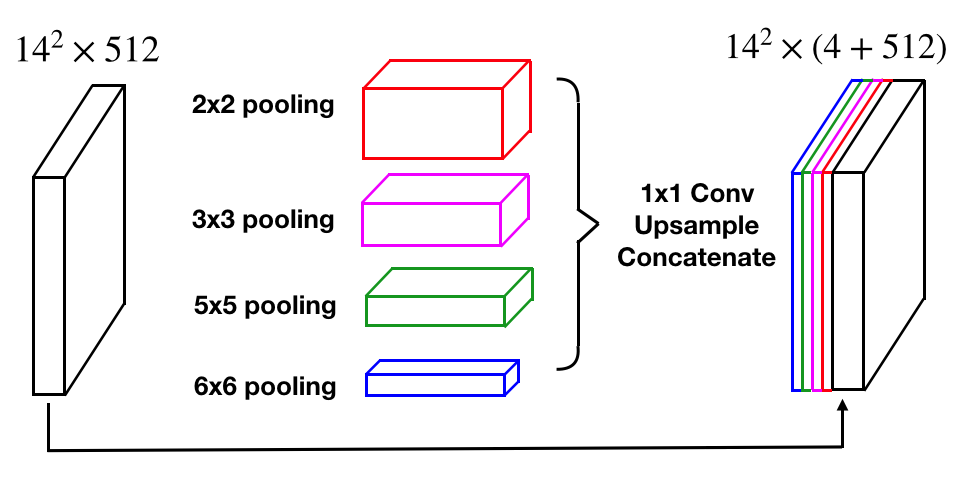}}
	\caption{The illustrations of residual multi-kernel pooling (RMP) strategy. The proposed RMP gather context information with four different-size pooling kernels. Then features are fed into 1$\times$1 convolution to reduce the dimension of feature maps. Finally, the upsampled features are concatenated with original features.}
	\label{picture3}
\end{figure}

3) \textbf{Residual Multi-kernel pooling}: A challenge in segmentation is the large variation of object size in medical image. For example, a tumor in middle or late stage can be much larger than that in early stage. In this paper, we propose a residual multi-kernel pooling to address the problem, which mainly relies on multiple effective field-of-views to detect objects at different sizes.

The size of receptive field roughly determines how much context information we can use. The general max pooling operation just employs a single pooling kernel, such as 2$\times$2. As illustrated in Fig.~\ref{picture3}, the proposed RMP encodes global context information with four different-size receptive fields: 2$\times$2, 3$\times$3, 5$\times$5 and 6$\times$6. The four-level outputs contain the feature maps with various sizes. To reduce the dimension of weights and computational cost, we use a 1$\times$1 convolution after each level of pooling. It reduces the dimension of the feature maps to the $\frac{1}{N}$ of original dimension, where $N$ represents number of channels in original feature maps. Then we upsample the low-dimension feature map to get the same size features as the original feature map via bilinear interpolation. Finally, we concatenate the original features with upsampled feature maps.

\subsection{Feature Decoder Module}

The feature decoder module is adopted to restore the high-level semantic features extracted from the feature encoder module and context extractor module. The skip connection takes some detailed information from the encoder to the decoder to remedy the information loss due to consecutive pooling and striding convolutional operations. Similar to \cite{apostolopoulos2017pathological}, we adopted an efficient block to enhance the decoding performance. The simple upscaling and deconvolution are two common operations of the decoder in the U-shape Networks. The upscaling operation increases the image size with linear interpolation, while deconvolution (also called transposed convolution) employs convolution operation to enlarge the image. Intuitively, the transposed convolution could learn a self-adaptive mapping to restore feature with more detailed information. Therefore, we choose to use the transposed convolution to restore the higher resolution feature in the decoder. As illustrated in Fig.~\ref{2}(c), it mainly includes a 1$\times$1 convolution, a 3$\times$3 transposed convolution and a 1$\times$1 convolution consecutively. Based on skip connection and the decoder block, the feature decoder module outputs a mask, the same size as the original input. 

\subsection{Loss Function}

Our framework is an end-to-end deep learning system. As illustrated in Fig.~\ref{2}, we need to train the proposed method to predict each pixel to be foreground or background, which is a pixel-wise classification problem. The most common loss function is cross entropy loss function.



However, the objects in medical images such as optic disc and retinal vessels often occupy a small region  in the image. The cross entropy loss is not optimal for such tasks. In this paper, we use the Dice coefficient loss function \cite{crum2006generalized, milletari2016v} to replace the common cross entropy loss. The comparison experiments and discussions are also conducted in the following section.  The Dice coefficient is a measure of overlap widely used to assess segmentation performance when ground truth is available, as in Equation (\ref{segmentation loss}):

\begin{equation}
    \label{segmentation loss}
L_{dice} = 1 - \sum_{k}^K\frac{2\omega_{k}\sum_{i}^{N}p_{(k,i)}g_{(k,i)}}{\sum_{i}^{N}p^{2}_{(k,i)} + \sum_{i}^{N}g^{2}_{(k,i)}}
\end{equation}
where $N$ is the pixel number, $p_{(k,i)} \in $  $\left[0,1\right]$ and $g_{(k,i)} \in $  $\left\{0,1\right\}$
denote predicted probability and ground truth label for class $k$, respectively. $K$ is the class number, and $\sum_{k}\omega_{k}$ = 1 are the class weights. In our paper, we set $\omega_{k} = \frac{1}{K}$ empirically.

  The final loss function is defined as:
\begin{equation}\label{total_equation}
L_{loss} = L_{dice} + L_{reg}
\end{equation}
where $L_{reg}$ represents the regularization loss (also called to weight decay) \cite{hoerl1970ridge} used to avoid overfitting.

To evaluate the performance of CE-Net, we apply the proposed method to five different medical image segmentation tasks: optic disc segmentation, retinal vessel detection, lung segmentation, cell contour segmentation and retinal OCT layer segmentation.

\section{Experiment}
\label{Experiment}
\subsection{Experimental Setup}
In this section, we first introduce the image preprocessing and data augmentation strategies   used in training and testing phases.

1) \textbf{Training phase}: 
Because of the limited number of training images, the datasets are augmented to reduce the risk of overfitting \cite{krizhevsky2012imagenet}. Firstly, we do data augmentation in an ambitious way, including horizontal flip, vertical flip and diagonal flip. In this way, each image in the original dataset is augmented to 2$\times$2$\times$2=8 images. Next,
the solutions of image preprocessing mainly include scaling from 90\% to 110\%, color jittering in HSV color space and image shifting randomly. The random image preprocessing method can enhance the data augmentation capability. 

2) \textbf{Testing phase}: To improve the robustness of medical image segmentation method, we also adopt test augmentation strategy, as that in \cite{dai2016r, zhang2018fully}, including image horizontal flip, vertical flip and diagonal flip (equal to predicting each image 8 times). Then we average the 8 predictions to get the final prediction map. All baseline approaches utilize the same strategy during testing phase.

3) \textbf{Experiment settings}:  
Our proposed network is based on the ResNet pretrained on ImageNet. The implementation is based on the public PyTorch platform. The training and testing bed is Ubuntu 16.04 system with the NVidia GeForce Titan graphics cards, which has 12 Gigabyte  memory.

During the  training, we adopt mini-batch stochastic gradient descent (SGD) with batch size 8, momentum 0.9 and weight decay 0.0001, other than Adam optimization. We use SGD optimization since recent studies \cite{wilson2017marginal} \cite{keskar2017improving} show that SGD often achieves a better performance, though the Adam optimization convergences faster. In addition, we use the “poly” learning rate policy where the learning rate is multiplied by  $(1-\frac{iter}{max\_iter})^{power}$ with power 0.9 and  
initial learning rate 4$e^{-3}$ \cite{zhao2017pyramid}. The maximum epoch is 100. We have released our codes on Github \footnote{https://github.com/Guzaiwang/CE-Net}.

\begin{table*}[t]
	\normalsize
\caption{Comparison with Different Methods for OD segmentation on the ORIGA, Messidor and RIM-ONE-R1 Datasets(mean$\pm$standard deviation)}
\centering

\begin{tabular}{c|c|c|c|c|c|c|c|c}
\hline
\multirow{2}{*}{Method} & \multirow{2}{*}{ORIGA} & \multirow{2}{*}{Messidor} & \multicolumn{6}{c}{RIM-ONE-R1}                                                                                                                                                              \\ \cline{4-9} 
                  &                        &                           & \multicolumn{1}{l|}{Expert 1} & \multicolumn{1}{l|}{Expert 2} & \multicolumn{1}{l|}{Expert 3} & \multicolumn{1}{l|}{Expert 4} & \multicolumn{1}{l|}{Expert 5} & \multicolumn{1}{c}{Overall} \\ \hline
Superpixel\cite{cheng2013superpixel}        & 0.102$\pm${0.104}                 & 0.125$\pm$0.113                     & 0.178                         & 0.229                         & 0.243                         & 0.183                         & 0.181                         & 0.203$\pm$0.104                        \\ \hline
U-Net \cite{ronneberger2015u}            & 0.115$\pm$0.068                & 0.069$\pm$0.121                     & 0.137                         & 0.149                         & 0.156                         & 0.171                         & 0.149                         & 0.152$\pm$0.107                        \\ \hline
M-Net\cite{fu2018joint}             & 0.071$\pm$0.047                 & 0.113$\pm$0.089                     & 0.128                         & 0.135                         & 0.153                         & 0.142                         & 0.117                         & 0.135$\pm$0.098                        \\ \hline
Faster RCNN \cite{jiang2018optic}       & 0.069$\pm$0.056                 &0.079$\pm$0.058                           & 0.101                              & 0.152                              & 0.161                              &  0.149                             &  0.104                             & 0.133$\pm$0.107        \\ \hline
DeepDisc \cite{gu2018deepdisc}         & 0.069$\pm$0.040                 & 0.064$\pm$0.039                     & 0.077                         & \textbf{0.107}                & \textbf{0.119}                & 0.101                         & 0.079                         &0.097$\pm$0.045                       \\ \hline
CE-Net              & \textbf{0.058$\pm$0.032}         & \textbf{0.051$\pm$0.033}            & \textbf{0.058}                & 0.112                         & 0.125                         & \textbf{0.080}                & \textbf{0.059}                & \textbf{0.087$\pm$0.039}               \\ \hline
\end{tabular}
\label{result-table}
\end{table*}

\subsection{Optic disc segmentation}
We first test the proposed CE-Net on optic disc segmentation. Three datasets, 
  ORIGA \cite{zhang2010origa}, Messidor \cite{decenciere2014feedback} and RIM-ONE-R1 \cite{fumero2011rim}, are used in our experiments. ORIGA dataset contains  650 images with dimension $3072\times 2048$. It has  been divided into 2 sets: \emph{Set A} for training and \emph{Set B} for testing \cite{cheng2018sparse}. In this paper, we follow the same partition of the data set to train and test our models. Messidor dataset is a public dataset provided by the Messidor program partners. It consists of 1200 images with three different sizes: 1440 $\times$ 960, 2240 $\times$ 1488, 2340 $\times$ 1536. The Messidor dataset is originally collected for Diabetic Retinopathy (DR) grading. Later, disc boundary for each image has also been provided from the official website \footnote{http://www.uhu.es/retinopathy/}. RIM-ONE dataset consists of three releases. The numbers of image are 169, 455 and 159 respectively. In this paper, we use first released dataset (RIM-ONE-R1), and  there are five different expert annotations in RIM-ONE-R1 dataset. We follow the partition in \cite{li2018learning} to get the training and testing images in the Messidor and RIM-ONE-R1 datasets. It should be noted that the ORIGA and Messidor datasets provide full image while the RIM-ONE-R1 provides cropped image.

In order to segment the optic disc in the retinal fundus images based on their original resolution, we crop an 800 $\times$ 800 area around the brightest point as motivated in \cite{zhang2010optic}, except for RIM-ONE-R1 dataset where the region with optic disc has already been cropped and provided.

To evaluate the performance, we adopt the overlapping error, which has been commonly used to evaluate the accuracy of optic disc segmentation:
\begin{equation}
E = 1- \frac{Area(S \cap G)}{Area(S \cup G)},
\end{equation}
where $S$ and $G$ denote the segmented and the manual ground truth optic disc respectively. Beside the average values, we also calculate the corresponding standard deviations.

\begin{figure*}[ht]
	\centering
	\includegraphics[width=0.95\linewidth]{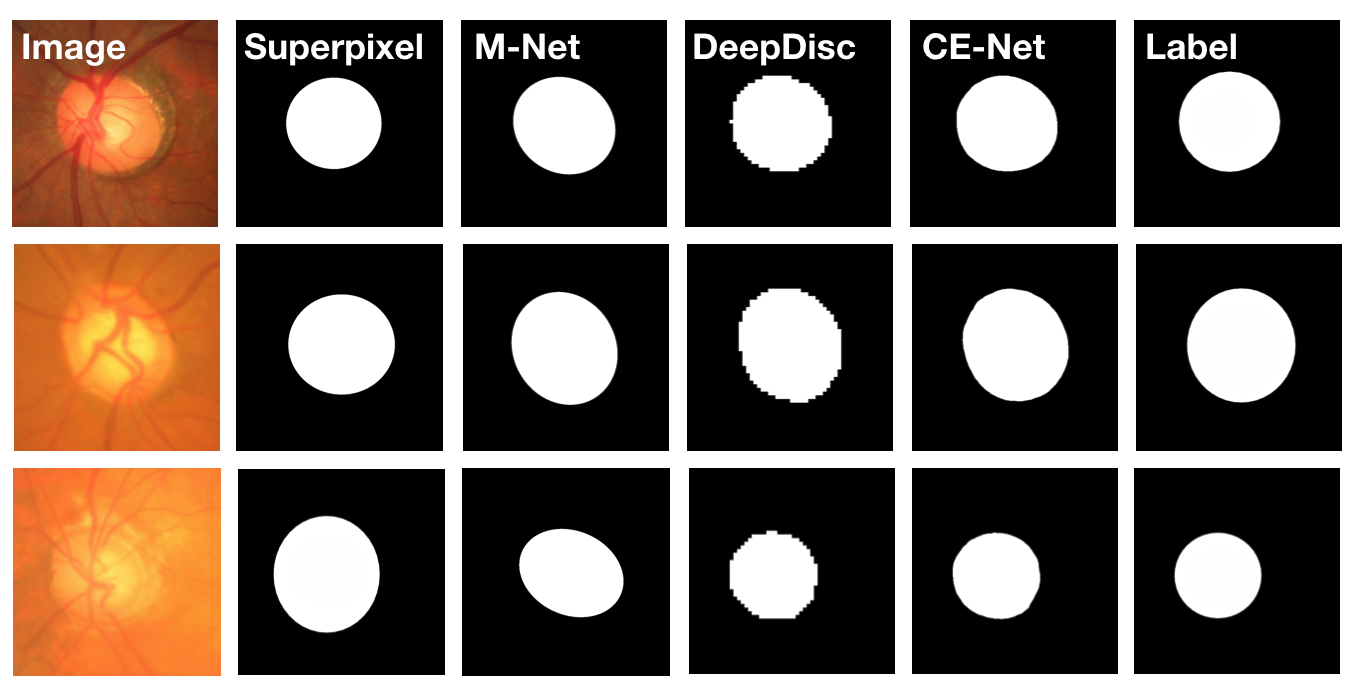}\vspace{-.12cm}
	\caption{Sample results. From left to right: original fundus images, state-of-the-art results obtained by superpixel based method \cite{cheng2013superpixel}, M-Net \cite{fu2018joint}, DeepDisc \cite{gu2018deepdisc}, CE-Net and ground-truth masks.}
	\label{6}\vspace{-.25cm}
\end{figure*}

We compare our methods with state-of-the-art algorithms. Five different algorithms are compared, including superpixel classification method \cite{cheng2013superpixel}, U-Net \cite{ronneberger2015u}, M-Net method \cite{fu2018joint}, faster RCNN  method \cite{ren2017faster} and DeepDisc method \cite{gu2018deepdisc}.  All of baseline models are adopted from their original implementations.

Table~\ref{result-table} shows the mean and standard deviation of the overlapping errors of these methods. As we can see, the proposed CE-Net outperforms the state-of-the-art optic disc segmentation methods.
In particular, it achieves an overlapping error of 0.058 in the ORIGA dataset, a relative reduction of 15.9\% from 0.069 by the latest Faster RCNN or DeepDisc methods. 
 In Messidor dataset, CE-Net achieves an overlapping error of 0.051, which is a relative reduction of  20.3\% from   0.064 by DeepDisc. The RIM-ONE-R1 dataset has five independent annotations. In our experiments, we follow the same setting in \cite{li2018learning} to use cross validation to get the results. Although it performs slightly worse than DeepDisc in comparison  with the annotation by Expert 2 and Expert 3, the overall results still show that CE-Net outperforms DeepDisc and other methods.  

We also show four sample results in Fig.~\ref{6} to visually compare our method with some competitive methods,  including superpixel based method, M-Net and DeepDisc. The images show that our method  obtain more accurate segmentation results. 

\begin{figure*}[ht]
	\centering
	\includegraphics[width=0.95\linewidth]{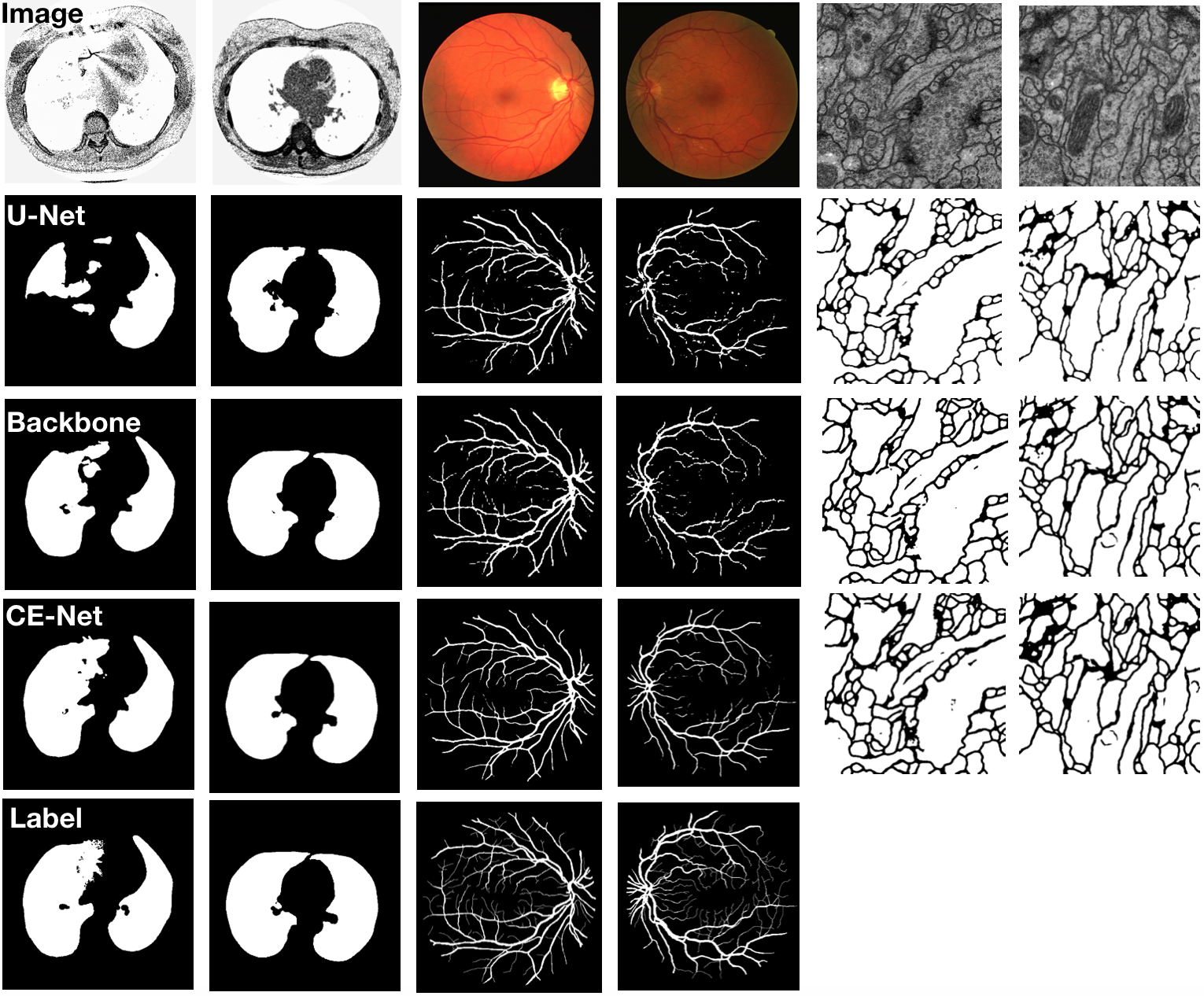}\vspace{-.11cm}
	\caption{Sample results of lung segmentation, vessel detection and cell contour segmentation. From top to bottom: original  images, U-Net, Backbone, CE-Net and ground truth (The ground truth for cell images is not given).  } 
	\label{all-comparison}\vspace{-.25cm}
\end{figure*}

\subsection{Retinal Vessel Detection}
The second application is the retinal vessel detection.
We use the public  DRIVE \cite{staal2004ridge} dataset which contains 40 images.  In DRIVE, two expert manual annotations are provided, the first of which is chosen as the ground truth for performance evaluation in the literature \cite{fu2016deepvessel}. The 40 images are divided into 20 images for training and 20 images for testing.   
To compare performance of the vessel detection, we compute two evaluation metrics, the sensitivity (Sen) and the accuracy (Acc), which are also calculated in \cite{fu2016deepvessel} \cite{liskowski2016segmenting}.

\begin{equation}
Sen = \frac{TP}{TP+FN}
\end{equation}

\begin{equation}
Acc = \frac{TP+TN}{TP+TN+FP+FN}
\end{equation}
where $TP$, $TN$, $FP$ and $FN$ represent the number of true positives, true negatives, false positives and false negatives, respectively. In addition, we also introduce the area under receiver operation characteristic curve (AUC)   to measure segmentation performance.

We compare the proposed CE-Net with the state-of-the-art algorithms \cite{zhao2015automated, azzopardi2015trainable, roychowdhury2015iterative}. In addition, some classical deep learning based methods \cite{xie2015holistically, ronneberger2015u, fu2016deepvessel} are also included into the comparison. 
Table~\ref{vessel} shows the comparison among these methods. 
From the comparison, the CE-Net achieves 0.8309, 0.9545  and 0.9779 in \textit{Sen}, \textit{Acc} and \textit{AUC} respectively, better than other methods. Comparing with the  backbone, the  \textit{Sen} increases from 0.7781 to 0.8309 by 6.8\%, the \textit{Acc} increases from 0.9477 to 0.9545 and  the \textit{AUC}  increases from 0.9705 to 0.9779, which shows that the proposed DAC and RMP blocks are beneficial for retina vessel detection as well. We show  some examples for visual comparsion  in Fig.~\ref{all-comparison}.
\begin{table}[h] 
	\normalsize
	\caption{Performance comparison of vessel detection} 
	\centering
	\begin{tabular}{c| c|c |c } \hline 
	Method & $Sen$ & $Acc$  & $AUC$ \\ \hline
	Azzopardi  \cite{azzopardi2015trainable} & 0.7655 & 0.9442& 0.9614\\ \hline
	Roychowdhury \cite{roychowdhury2015iterative}  & 0.7250 & 0.9520 &0.9672 \\ \hline
	zhao \cite{zhao2015automated} & 0.7420 & 0.9540 & 0.8620 \\ \hline
	HED \cite{xie2015holistically} & 0.7364 & 0.9434  & 0.9723 \\  \hline
	U-Net \cite{ronneberger2015u} & 0.7537 &  0.9531 &  0.9601\\ \hline
	DeepVessel \cite{fu2016deepvessel} & 0.7603  & 0.9523 & 0.9752\\ \hline
	Backbone & 0.7781  & 0.9477 & 0.9705\\ \hline
	{CE-Net}  & \textbf{0.8309} & \textbf{0.9545} & \textbf{0.9779} \\ \hline
	\end{tabular}
	\label{vessel}
\end{table}

\subsection{Lung segmentation}
The next application is lung segmentation task, which is to segment lung structure in 2D CT images from the Lung Nodule Analysis (LUNA) competition. The LUNA competition is originally conducted for the following challenge tracks: nodule detection and false positive reduction.
Because the segmented lungs are fundamental for further lung nodule candidates, we adopt the challenge dataset to evaluate our proposed CE-Net. The dataset contains 534 2D samples (512$\times$512 pixels) with respective label images and can be freely downloaded from the official website \footnote{https://www.kaggle.com/kmader/finding-lungs-in-ct-data/data/}. We use 80\% of the images for training and the rest for testing, and cross validation is also conducted. The evaluation metrics include the overlapping error, accuracy and sensitivity, similar to those in optic disc segmentation and vessel detection. Beside the average values, we also calculate the corresponding standard deviations in Table \ref{lung}.

From the comparison shown in Table~\ref{lung}, the CE-Net achieves 0.038 in overlapping error, 0.8309 in Sensitivity score and 0.9545 in Accuracy score, better than the U-Net. We also compare CE-Net with the backbone, and the overlapping error decreases from 0.044 to 0.038 by 13.6\%, the sensitivity score increases from 0.967 to 0.980 while the accuracy increases from 0.988 to 0.990, which further supports that our proposed DAC and RMP blocks are beneficial for lung segmentation. We also give a few examples for visual comparison of lung segmentation in Fig.~\ref{all-comparison}.

\begin{table}[h] 
	\normalsize
	\caption{Performance comparison of lung segmentation(mean$\pm$standard deviation)} 
	\centering
\begin{tabular}{c|c|c|c}
\hline
Method   & $E$ & $Acc$ & $Sen$ \\ \hline
U-Net \cite{ronneberger2015u}   & 0.087$\pm$0.090  &  0.975$\pm$0.032   &  0.938   \\ \hline
Backbone & 0.044$\pm$0.063  &  0.988$\pm$0.024   &  0.967  \\ \hline
{CE-Net}   & \textbf{0.038$\pm$0.061}  & \textbf{0.990$\pm$0.023}    &  \textbf{0.980}   \\ \hline
\end{tabular}
	\label{lung}
\end{table}

\subsection{Cell contour segmentation}
The fourth application is cell contour segmentation.
  The cell segmentation task is to segment neuronal structures in electron microscopic recordings. The dataset is provided by the EM challenge, which started at ISBI 2012 and is still open for new contributions \cite{cardona2010integrated}. The training set contains 30 images (512$\times$512 pixels), and could be downloaded from the official website\footnote{http://brainiac2.mit.edu/}. 
The testing set consists of 30 images, and is publicly available as well. However, the corresponding ground truths are kept unknown. The results on the testing set are obtained by  sending the prediction maps to the organizers, who will then compute and release the results. From the statement on the official website, the following metrics are the best for the quantitative evaluation of segmentation results: foreground-restricted rand scoring after border thinning ($V^{Rand}$) and foreground-restricted information theoretic scoring after border thinning ($V^{Info}$). The $V^{Rand}$ mainly computes the weighted harmonic mean by jointing the Rand split score and Rand merge score, which are used to measure the segmentation performance. Similarly, the $V^{Info}$ mainly computes weighted harmonic mean of information theoretic score. The higher scores represent the better segmentation performance. The specific computation process and more details of these two algorithms could be found in \cite{arganda2015crowdsourcing}.

We compare our CE-Net with the original U-Net and backbone, and the final results are shown in Table~\ref{cell}. Our CE-Net outperforms the U-Net and Backbone. It indicates that our proposed CE-Net is effective for cell contour segmentation task. We also give a few examples for visual comparison  in Fig.~\ref{all-comparison}, though the ground truth is not available.

\begin{table}[h] 
	\normalsize
	\caption{Performance comparison of cell contour segmentation} 
	\centering
\begin{tabular}{c|l|l}  \hline
Method  & \multicolumn{1}{c|}{$V^{Rand}$} & \multicolumn{1}{c}{$V^{Info}$}                \\ \hline
U-Net \cite{ronneberger2015u}   & 0.9432                         & 0.9562                         \\ \hline
Backbone & 0.9569                         & 0.9716                                                     \\ \hline
CE-Net   & \textbf{0.9743}                & \textbf{0.9878}                                  \\ \hline
\end{tabular}
	\label{cell}
\end{table}

\begin{table*}[t]
\centering
\caption{The comparison results on Topcon dataset}\label{Topcon-label}
\begin{tabular}{c|c|c|c|c|c|c|c|c|c|c|c}
\hline
 Method& ILM   & NFL/GCL & IPL/INL & INL/OPL & OPL/ONL & ELM   & Up IS/OS & Low IS/OS & OS/RPE & BM/Choroid & Overall \\ \hline
  Topcon \cite{cheng2016speckle} & 1.61 & 2.09 & 2.10 & 2.27 & - & 2.17 & 1.82 & - & 1.65 &1.80  &-\\ \hline
  SRR \cite{cheng2016speckle} & 1.61 & \textbf{2.02} & \textbf{2.02} & 1.91 & - & 1.86 & 1.63 & - & 1.62 & 1.80 & -\\ \hline
  FCN \cite{long2015fully} & 2.10 & 4.41 & 3.77 & 4.54 & 4.78 & 4.52 & 3.84 & 4.36 & 5.06 & 7.88  &4.53\\ \hline
  U-Net \cite{ronneberger2015u} & 1.38 & 3.05 & 2.70 & 2.77 & 3.30 & 2.34 & 1.86 & 2.00 & 2.42 & 2.65 & 2.45 \\ \hline
Backbone & 2.13 & 2.70 & 2.52 & 2.20 & 2.79 & 1.91 & 1.26 & 1.60 & 2.02 & 2.70 &2.18 \\ \hline
CE-Net w/ CE & 1.45 & 2.48 & 2.20 & 2.08 & 2.55 &1.66 & 1.19 & \textbf{1.04} & 1.52 & 1.82 & 1.80\\ \hline
  CE-Net w/ Dice & \textbf{1.37} & \textbf{2.02} & 2.08 & \textbf{1.80} & \textbf{2.47} & \textbf{1.48} & \textbf{1.10} & 1.26 & \textbf{1.48} & \textbf{1.74} & \textbf{1.68}\\ \hline

\end{tabular}
\end{table*}

\begin{figure*}[t]
    \begin{center}

    \includegraphics[width=18cm]{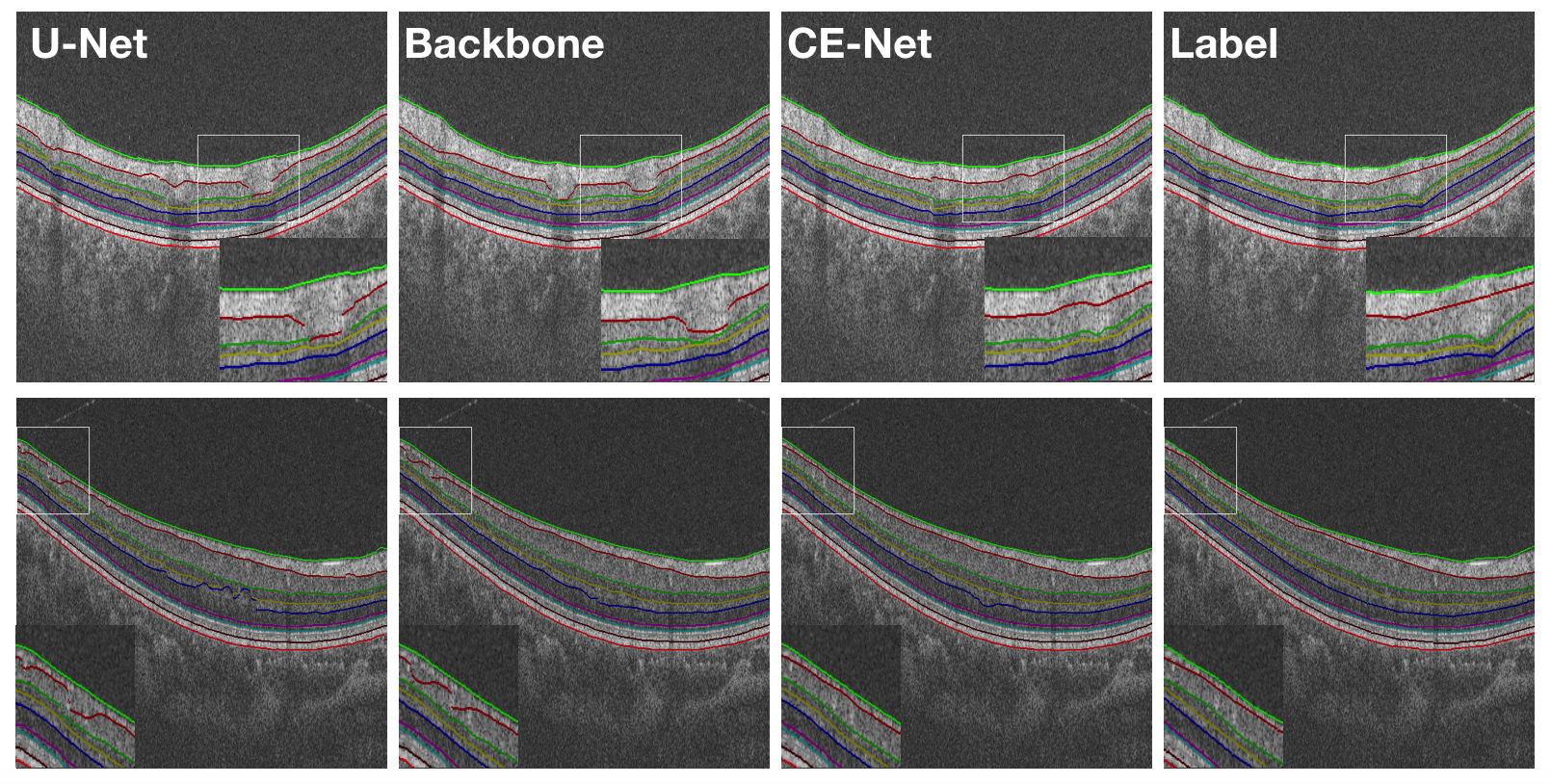}
    \end{center}
       \caption{Sample results. From left to right: U-Net, Backbone, CE-Net and ground-truth masks. The edges between different layers have been marked with colored lines}
    \label{topcon-demo1}
\end{figure*}

\subsection{Retinal OCT Layer Segmentation}
The above four application are conducted on two-class segmentation problems where we only need to segment foreground objects from background. In this paper, we also show that our method is applicable for multi-class segmentation tasks. We use the retinal OCT layer segmentation as an example to apply CE-Net to segment 11 retinal layers \cite{cheng2016speckle}. This dataset contains 20 3D volumes and each volume has 256 2D scans. Ten boundaries have been manually demarcated to divide each 2D image into 11 parts: boundary 1 corresponding to internal limiting membrane(ILM); boundary 2 between nerve fiber layer and the ganglion cells layer (NFL/GCL); boundary 3 between inner plexiform layer and the inner nuclear layer (IPL/INL); boundary 4 between the inner nuclear layer and the outer plexiform layer (INL/OPL); boundary 5 between the outer plexiform layer and the outer nuclear layer (OPL/ONL); boundary 6 corresponding to the external limiting membrane (ELM); boundary 7 corresponding to the upper boundary of inner segment (up IS); boundary 8 corresponding to the lower boundary of inner segment (low IS); boundary 9 between the outer segments and the retinal pigment epithelium (OS/RPE); boundary 10 between Bruch’s membrane and the choroid (BM/Choroid). 
To evaluate the performance, we adopt the mean absolute error \cite{cheng2016speckle}, which has been commonly used to evaluate the accuracy of retinal OCT layer segmentation.

We compare our proposed method with some state-of-the-art OCT layer segmentation approaches: Topcon built-in method in \cite{cheng2016speckle}, Speckle Reduction by Reconstruction (SRR) method \cite{cheng2016speckle}, FCN\cite{long2015fully} and U-Net\cite{ronneberger2015u}. 

The performance comparisons are summarized in Table \ref{Topcon-label}. Compared with U-Net and the backbone approach, our CE-Net achieves an overall mean absolute error of 1.68, which is a relative reduction of 31.4\% from 2.45 and 22.9\% from the 2.18, respectively. Compared to the Topcon built-in method and SRR, our CE-Net also achieves better results in most scenarios. This indicates that our proposed CE-Net could also be applied to multi-class segmentation tasks. Further, we also conduct the comparison experiments between cross entropy loss and dice loss. Table \ref{Topcon-label} shows that the CE-Net with dice loss is superior to that with cross entropy loss. 

We also present some sample results in Fig.~\ref{topcon-demo1} to visually compare our method with U-Net and the Backbone approach. The images clearly show more accurate segmentation results by our CE-Net. 

\subsection{Ablation Study }
 
 \begin{table*}[h] 
	\normalsize
	\caption{Ablation study for each component on ORIGA and DRIVE datasets} 
	\centering
\begin{tabular}{c|c|c|c}
\hline
\multirow{2}{*}{Method} & \multirow{1}{*}{ORIGA} & \multicolumn{2}{c}{DRIVE}                 \\ \cline{2-4} 
 & \multicolumn{1}{c|}{$E$} & \multicolumn{1}{c|}{$Acc$} & \multicolumn{1}{c}{$AUC$} \\ \hline
	U-Net  & 0.115$\pm$0.068 & 0.939$\pm$0.006& 0.960$\pm$0.006 \\ \hline
	Backbone  & 0.075$\pm$0.068 & 0.943$\pm$0.004& 0.971$\pm$0.005\\ \hline
	Backbone + Inception-block & 0.068$\pm$0.059& 0.950$\pm$0.004& 0.972$\pm$0.005 \\ \hline
    Backbone + DAC w/o atrous & 0.073$\pm$0.050 & 0.952$\pm$0.004& 0.970$\pm$0.005 \\ \hline
	Backbone + DAC with atrous & 0.061$\pm$0.043 & 0.953$\pm$0.004& 0.977$\pm$0.006 \\ \hline
	Backbone + RMP & 0.061$\pm$0.044 & 0.952$\pm$0.004& 0.974$\pm$0.005 \\ \hline
	Backbone + Inception-ResNet-block & 0.065$\pm$0.042 & 0.951$\pm$0.004& 0.974$\pm$0.005\\ \hline

	CE-Net & \textbf{0.058$\pm$0.032} & \textbf{0.955$\pm$0.003} & \textbf{0.978$\pm$0.006} \\ \hline
	\end{tabular}
	\label{ablation_study}
\end{table*}

\begin{figure}[h]
	\centering
	\subfigure{
	  \includegraphics[width=8cm]{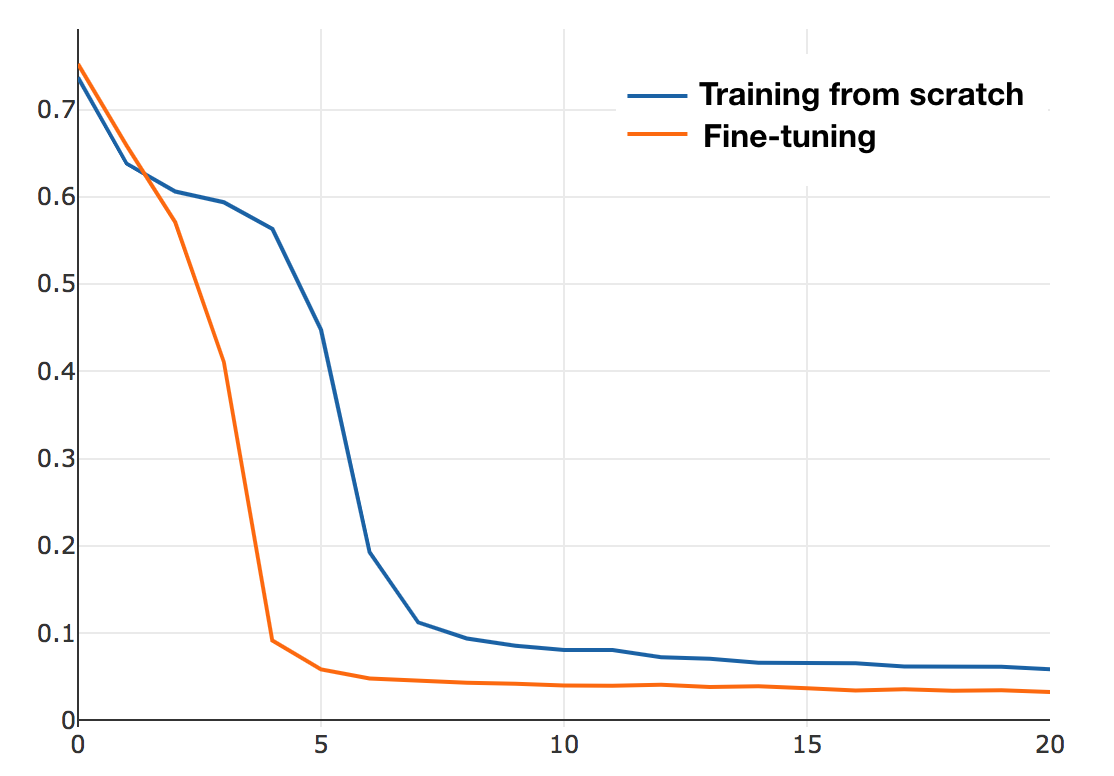}}
	\caption{The orange line represents the training loss of fine-tuning from pretrained weights, while the blue represents loss of end-to-end training from scratch.}
	\label{rebuttal-loss}
\end{figure}

In order to justify the effectiveness of the pretrained ResNet, DAC block and RMP block in the proposed CE-Net, we conduct the following ablation studies using the ORIGA and DRIVE datasets as examples: 

\textbf{Ablation study for adopting pretrained ResNet model:} Our proposed method is based on U-Net, therefore U-Net is the most fundamental baseline model. We employed the residual block to replace the original encoder block of U-Net, aiming at enhancing the learning capability. We call the modified U-shape network with pretrained residual block and feature decoder as `Backbone'. Recent work \cite{he2018rethinking} points out that ImageNet pre-training largely helps to circumvent optimization problems and fine-tuning from pretrained
weights converges faster than that from scratch. We have also conducted experiments to compare the results with pre-training to those without. Fig.~\ref{rebuttal-loss} shows how the losses change in the two scenarios. As we can see, the loss decreases faster in the case with pretraining than that without. Table~\ref{ablation_study} shows the  segmentation results. By adopting the pretrained ResNet blocks, the Backbone approach achieved a better   performance. For OD segmentation, the overlapping error is decreased by 34.8\% from 0.115 to 0.075. For retinal vessel detection, the \textit{Acc}  and \textit{AUC} are increased from 0.939 and 0.960 to 0.943 and 0.971, respectively. The results  indicate  that pretrained ResNet blocks are beneficial.

\textbf{Ablation study for dense atrous convolution block:} The proposed DAC block employs the atrous convolution with different rates, assembled in the Inception-like block. Therefore, we first conduct experiments to validate the usefulness of the atrous convolution. We use regular convolution to replace the atrous convolution in DAC block (referred to Backbone + DAC w/o atrous). As shown in Table~\ref{ablation_study}, our proposed DAC module (referred to Backbone + DAC with atrous) reduces the overlapping error by 16.4\% from 0.073 to 0.061 in OD segmentation and improves  the \textit{Acc}  and \textit{AUC}  in retinal vessel detection. This indicates that atrous convolution helps to extract high-level semantic features, compared to the regular convolution. We also compare our proposed DAC block with the regular Inception-V2 block (referred to Backbone + Inception-block). The comparison results show that the DAC block outperforms the regular inception block, with a relative reduction of 10.3\% from 0.068 to 0.061 in overlapping error for OD segmentation.
Finally, the overlapping error is reduced by 18.7\% from 0.075 of Backbone to 0.061 (Backbone + DAC). This shows that the proposed DAC block is able to further extract global information to get high-level semantic feature maps with high resolution, which is useful for our segmentation task.  

\textbf{Ablation study for residual multi-kernel pooling module:} Table~\ref{ablation_study} also shows the effect of RMP, which boosts the performance of OD segmentation. The Backbone with RMP module is referred to as `Backbone + RMP'. Compared to the Backbone, the overlapping error decreased by 18.7\% from 0.075 to 0.061 in OD segmentation, while the  \textit{Acc} and \textit{AUC} scores increased from 0.943 and 0.971 to 0.952 and 0.974 for retinal vessel detection. The RMP module could encode the global information and change the combination way of feature maps.

\textbf{Ablation study for network with similar complexity:} Researchers have shown that the complexity is an embodiment of the network capability \cite{huang2017densely} and an increased complexity often leads to better performance. Therefore, there is a concern that the improvements might come from the increased complexity of the network.    To ease such a concern, we compare our network with a network with similar complexity. In this paper, we compare it with  the aforementioned backbone backed up by regular Inception-ResNet-V2 blocks (Backbone + Inception-ResNet-block).  Table~\ref{ablation_study} shows that our CE-Net is better, with an overlapping error reduction from 0.065 to 0.058 in OD segmentation and the \textit{Acc} and \textit{AUC} scores increase from 0.951 and 0.974 to 0.955 and 0.978.

\section{CONCLUSIONS}
\label{Conclusion}
Medical image segmentation is important in the medical image analysis. 
In this paper, we propose an end-to-end deep learning framework named CE-Net for medical image segmentation. Compared with U-Net, the proposed CE-Net adopts pretrained ResNet block in the feature encoder. A newly proposed dense atrous convolution block and residual multi-kernel pooling are integrated to the ResNet modified U-Net structure to capture more high-level features and preserve more spatial information. Our method can be applied to a new application by fine-tuning our model using the new training data and the manual ground truth. Our experimental results show that the proposed method is able to improve the medical image segmentation in different tasks, including optic disc segmentation, retinal vessel detection, lung segmentation, cell contour segmentation and retinal OCT layer segmentation. It is believed that the approach is a general one and can be applied to other 2D medical  image segmentation tasks. In this paper, our method is validated on 2D images now and the extension to 3D data would be a possible future work.

\bibliographystyle{IEEEtran}

\bibliography{IEEEabrv}

\end{document}